\pdfoutput=1

\documentclass[11pt]{article}
\usepackage{authblk}
\usepackage{ACL2023}

\usepackage{times}
\usepackage{latexsym}

\usepackage[T1]{fontenc}

\usepackage[utf8]{inputenc}

\usepackage{microtype}

\usepackage{inconsolata}
\usepackage{subfigure}

\usepackage{amsmath}
\usepackage{amssymb}
\usepackage{mathtools}
\usepackage{amsthm}
\usepackage{colortbl}
\usepackage{multirow}

\title{\texttt{ORPO}: Monolithic Preference Optimization without Reference Model} 

\author{\textbf{Jiwoo Hong}}
\author{\textbf{Noah Lee}}
\author{\textbf{James Thorne}}

\affil{KAIST AI \protect \\ 
    \tt\{jiwoo\_hong, noah.lee, thorne\}@kaist.ac.kr}

\begin{document}

\maketitle

\begin{abstract}
While recent preference alignment algorithms for language models have demonstrated promising results, supervised fine-tuning (SFT) remains imperative for achieving successful convergence. In this paper, we study the crucial role of SFT within the context of preference alignment, emphasizing that a minor penalty for the disfavored generation style is sufficient for preference-aligned SFT. Building on this foundation, we introduce a straightforward and innovative reference model-free monolithic odds ratio preference optimization algorithm, \texttt{ORPO}, eliminating the necessity for an additional preference alignment phase. We demonstrate, both empirically and theoretically, that the odds ratio is a sensible choice for contrasting favored and disfavored styles during SFT across the diverse sizes from 125M to 7B. Specifically, fine-tuning Phi-2 (2.7B), Llama-2 (7B), and Mistral (7B) with \texttt{ORPO} on the UltraFeedback alone surpasses the performance of state-of-the-art language models with more than 7B and 13B parameters: achieving up to 12.20\% on $\text{AlpacaEval}_{2.0}$ (Figure \ref{fig:first}), 66.19\% on IFEval (instruction-level loose, Table \ref{tab:ifeval}), and 7.32 in MT-Bench (Figure \ref{fig:mtbench}). We release code\footnote{\url{https://github.com/xfactlab/orpo}} and model checkpoints for Mistral-\texttt{ORPO}-$\alpha$ (7B)\footnote{\url{https://huggingface.co/kaist-ai/mistral-orpo-alpha}} and Mistral-\texttt{ORPO}-$\beta$ (7B).\footnote{\url{https://huggingface.co/kaist-ai/mistral-orpo-beta}}
\end{abstract}

\setlength{\jot}{10pt}
\section{Introduction}
Pre-trained language models (PLMs) with vast training corpora such as web texts \citep{Gokaslan2019OpenWeb, penedo2023refinedweb} or textbooks \citep{li2023textbooks} have shown remarkable abilities in diverse natural language processing (NLP) tasks \citep{NEURIPS2020_1457c0d6, zhang2022opt, touvron2023llama, jiang2023mistral, almazrouei2023falcon}. 
\begin{figure}[hbt]
    \centering
    \includegraphics[width=\columnwidth]{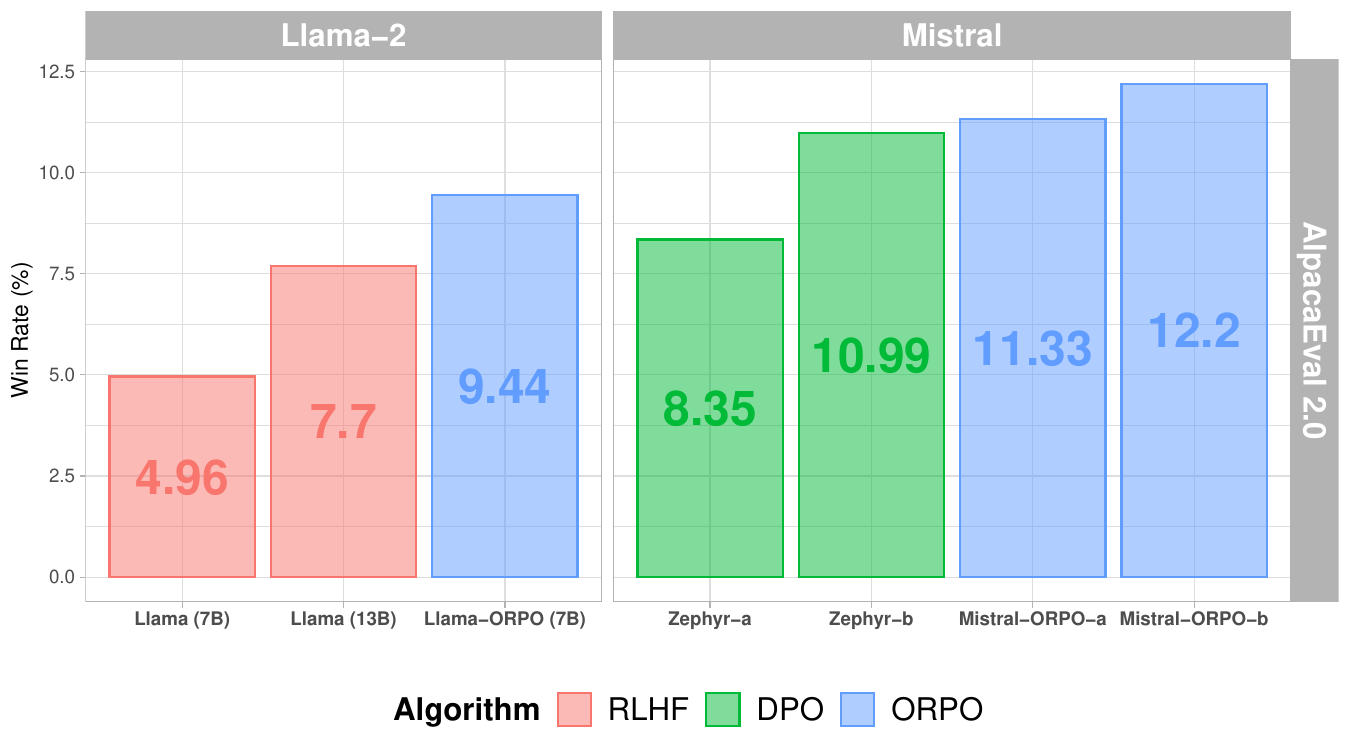}
    \caption{$\text{AlpacaEval}_{2.0}$ result of Llama-2 (7B) and Mistral (7B) fine-tuned with \texttt{ORPO} (blue) in comparison to the state-of-the-art models. Notably, Mistral-\texttt{ORPO}-$\alpha$ \& $\beta$ surpasses Zephyr $\beta$ and Llama-2-Chat (13B) with a single epoch training exclusively on the UltraFeedback.}
    \label{fig:first}
\end{figure}However, the models must undergo further tuning to be usable in general-domain applications, typically through processes such as \textit{instruction tuning} and \textit{preference alignment}. 

Instruction-tuning \citep{wei2022finetuned, alpaca, wang2023far, zhou2023lima} trains models to follow task descriptions given in natural language, which enables models to generalize well to previously unseen tasks. However, despite the ability to follow instructions, models may generate harmful or unethical outputs \citep{carlini2021extracting, gehman-etal-2020-realtoxicityprompts, pryzant-etal-2023-automatic}.  To further align these models with human values, additional training is required with pairwise preference data using techniques such as reinforcement learning with human feedback \citep[RLHF]{ziegler2020finetuning, stiennon2022learning} and direct preference optimization \citep[DPO]{rafailov2023direct}.

\begin{figure*}[hbt!]
    \centering
    \includegraphics[width=\textwidth]{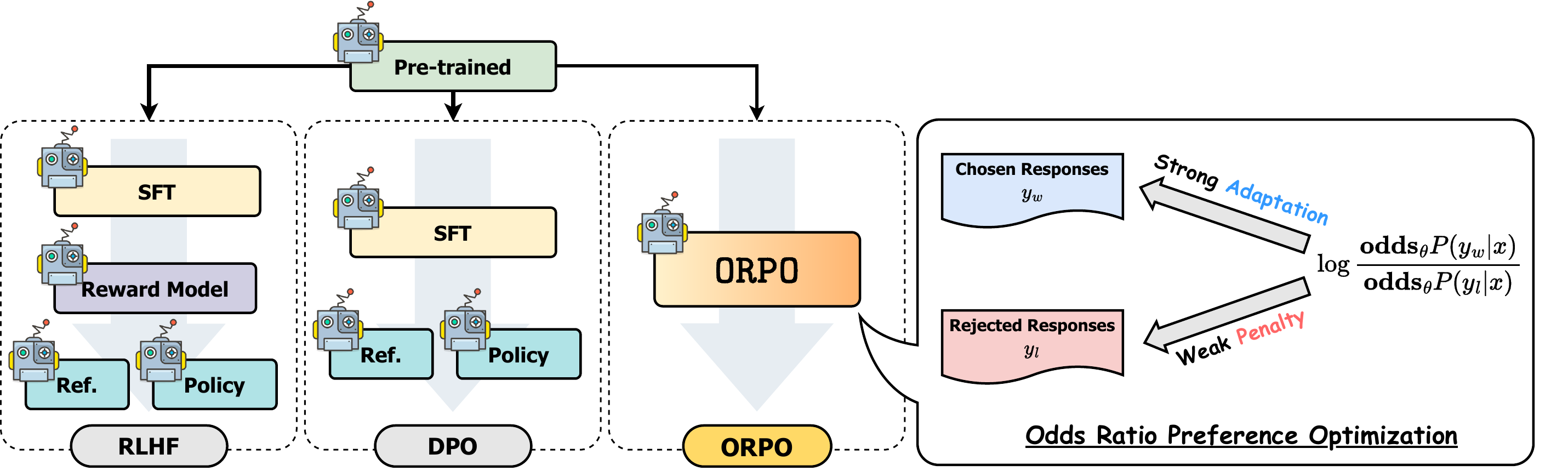}
    \caption{Comparison of model alignment techniques. \texttt{ORPO} aligns the language model \textit{without a reference model} in a single-step manner by assigning a weak penalty to the rejected responses and a strong adaptation signal to the chosen responses with a simple log odds ratio term appended to the negative log-likelihood loss.}
    \label{fig:main}
\end{figure*}

Preference alignment methods have demonstrated success in several downstream tasks beyond reducing harm. For example, improving factuality \citep{tian2023finetuning, cheng2024ai, chen2024grath}, code-based question answering \citep{gorbatovski2024reinforcement}, and machine translation \citep{ramos2023aligning}. The versatility of alignment algorithms over a wide range of downstream tasks highlights the necessity of understanding the alignment procedure and further improving the algorithms in terms of efficiency and performance. However, existing preference alignment methods normally consist of a multi-stage process, as shown in Figure \ref{fig:main}, typically requiring a second reference model and a separate warm-up phase with supervised fine-tuning (SFT) \citep{ziegler2020finetuning, rafailov2023direct, wu2023pairwise}. 

In this paper, we study the role and impact of SFT in pairwise preference datasets for model alignment in Section \ref{sec:role} and propose a simple and novel monolithic alignment method, odds ratio preference optimization (\texttt{ORPO}), which efficiently penalizes the model from learning undesired generation styles during SFT in Section \ref{sec:method}. In contrast to previous works, our approach requires neither an SFT warm-up stage nor a reference model, enabling resource-efficient development of preference-based aligned models. 

We demonstrate the effectiveness of our method with the evaluation of model alignment tasks and popular leaderboards in Section \ref{subsec:instruct} and \ref{subsec:multi} by fine-tuning Phi-2 (2.7B), Llama-2 (7B), and Mistral (7B) with \texttt{ORPO}. Then, we conduct controlled experiments comparing \texttt{ORPO} against established methods for model alignment, RLHF, and DPO for different datasets and model sizes in Section \ref{subsec:rmwin}. Along with the post-hoc analysis of generation diversity in Section \ref{subsec:div}, we expound on the theoretical, empirical, and computational justification of utilizing the odds ratio in monolithic preference alignment in Section \ref{subsec:compute}. We release the training code and the checkpoints for Mistral-\texttt{ORPO}-$\alpha$ (7B) and Mistral-\texttt{ORPO}-$\beta$ (7B). These models achieve 7.24 and 7.32 in MT-Bench, 11.33\% and 12.20\% on $\text{AlpacaEval}_{2.0}$, and 61.63\% and 66.19\% in IFEval instruction-level loose accuracy, respectively. 

\section{Related Works}
\paragraph{Alignment with Reinforcement Learning} Reinforcement learning with human feedback (RLHF) commonly applies the Bradley-Terry model \citep{19ff28b9-64f9-3656-ba40-08326a05748e} to estimate the probability of a pairwise competition between two independently evaluated instances. An additional reward model is trained to score instances. Reinforcement learning algorithms such as proximal policy optimization (PPO) \citep{schulman2017proximal} are employed to train the model to maximize the score of the reward model for the chosen response, resulting in language models that are trained with human preferences \citep{ziegler2020finetuning, stiennon2022learning}. Notably, \citet{ouyang2022training} demonstrated the scalability and versatility of RLHF for instruction-following language models. Extensions such as language model feedback (RLAIF) could be a viable alternative to human feedback \citep{bai2022constitutional, lee2023rlaif, pang2023language}. However, RLHF faces challenges of extensive hyperparameter searching due to the instability of PPO \citep{rafailov2023direct, wu2023pairwise} and the sensitivity of the reward models \citep{gao2022scaling, wang2024secrets}. Therefore, there is a crucial need for stable preference alignment algorithms.

\paragraph{Alignment without Reward Model} Several techniques for preference alignment mitigate the need for reinforcement learning \citep{rafailov2023direct, song2023preference, azar2023general, ethayarajh2023halos}. \citet{rafailov2023direct} introduce direct policy optimization (DPO), which combines the reward modeling stage into the preference learning stage. \citet{azar2023general} prevented potential overfitting problems in DPO through identity preference optimization (IPO). \citet{ethayarajh2023halos} and \citet{Cai2023ULMAUL} proposed Kahneman-Tversky Optimisation (KTO) and Unified Language Model Alignment (ULMA) that does not require the pair-wise preference dataset, unlike RLHF and DPO. \citet{song2023preference} further suggests incorporation of the softmax value of the reference response set in the negative log-likelihood loss to merge the supervised fine-tuning and preference alignment. 

\paragraph{Alignment with Supervised Fine-tuning} Preference alignment methods in reinforcement learning (RL) often leverage supervised fine-tuning (SFT) to ensure the stable update of the active policy in relation to the old policy \citep{schulman2017proximal}. This is because the SFT model is the old policy in the context of RLHF \citep{ziegler2020finetuning}. Furthermore, empirical findings indicate that, even in non-RL alignment methods, the SFT model is crucial for achieving convergence to desired results \citep{rafailov2023direct, tunstall2023zephyr}. 

In contrast, there have been approaches to build human-aligned language models by conducting SFT only with filtered datasets \citep{zhou2023lima, li2023selfalignment, haggerty2024selfsupervised, Zhou2023LoBaSSGL}. \citet{zhou2023lima} demonstrated that SFT with a small amount of data with fine-grained filtering and curation could be sufficient for building helpful language model assistants. Furthermore, \citet{li2023selfalignment} and \citet{haggerty2024selfsupervised} proposed an iterative process of fine-tuning the supervised fine-tuned language models with their own generations after fine-grained selection of aligned generations and \citet{Zhou2023LoBaSSGL} suggested that a curated subset of preference dataset is sufficient for alignment. While these works highlight the impact and significance of SFT in the context of alignment, the actual role of SFT and the theoretical background for incorporating preference alignment in SFT remains understudied.

\section{The Role of Supervised Fine-tuning}\label{sec:role}
\begin{figure}[t]
    \centering
    \includegraphics[width=\columnwidth]{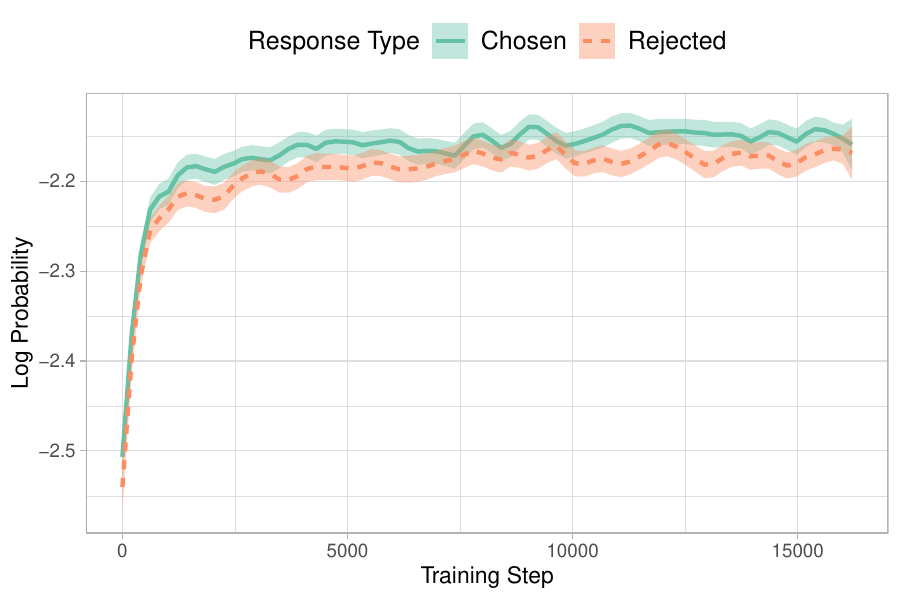}
    \caption{Log probabilities for chosen and rejected responses during OPT-350M model fine-tuning on HH-RLHF dataset. Despite only chosen responses being used for supervision, rejected responses show a comparable likelihood of generation.}
    \label{fig:sft}
\end{figure}

We study the behavior of supervised fine-tuning (SFT) as an initial stage of preference alignment methods \citep{ziegler2020finetuning, rafailov2023direct} through analysis of the loss function in SFT and empirical demonstration of the preference comprehension ability of the trained SFT model. SFT plays a significant role in tailoring the pre-trained language models to the desired domain \citep{zhou2023lima,dong2024abilities} by increasing the log probabilities of pertinent tokens. Nevertheless, this inadvertently increases the likelihood of generating tokens in undesirable styles, as illustrated in Figure \ref{fig:sft}. Therefore, it is necessary to develop methods capable of preserving the domain adaptation role of SFT while concurrently discerning and mitigating unwanted generation styles.

\paragraph{Absence of Penalty in Cross-Entropy Loss} The goal of cross-entropy loss model fine-tuning is to penalize the model if the predicted logits for the reference answers are low, as shown in Equation \ref{eq:celoss}. 
\begin{align}
\mathcal{L} &= -\frac{1}{m} \sum_{k=1}^{m} \log P(\mathbf{x}^{(k)}, \mathbf{y}^{(k)})\\
&= -\frac{1}{m}\sum_{k=1}^{m}\sum_{i=1}^{|V|}y_i^{(k)} \cdot \log(p_i^{(k)})\label{eq:celoss}
\end{align}
where $y_i$ is a boolean value that indicates if $i$th token in the vocabulary set $V$ is a label token, $p_i$ refers to the probability of $i$th token, and $m$ is the length of sequence. Using cross-entropy alone gives no direct penalty or compensation for the logits of non-answer tokens \citep{lin2017focal} as  $y_i$ will be set to 0. While cross-entropy is generally effective for domain adaptation  \citep{mao2023crossentropy}, there are no mechanisms to penalize rejected responses when compensating for the chosen responses. Therefore, the log probabilities of the tokens in the rejected responses increase along with the chosen responses, which is not desired from the viewpoint of preference alignment.

\paragraph{Generalization over Both Response Styles} We conduct a pilot study to empirically demonstrate the miscalibration of chosen and rejected responses with supervised fine-tuning alone. We fine-tune OPT-350M \citep{zhang2022opt} on \textit{the chosen responses only} from the HH-RLHF dataset \citep{bai2022constitutional}. Throughout the training, we monitor the log probability of rejected responses for each batch and report this in Figure \ref{fig:sft}. Both the log probability of chosen and rejected responses exhibited a simultaneous increase. This can be interpreted from two different perspectives. First, the cross-entropy loss effectively guides the model toward the intended domain (e.g., dialogue). However, the absence of a penalty for unwanted generations results in rejected responses sometimes having even higher log probabilities than the chosen ones.

\paragraph{Penalizing Undesired Generations} Appending an unlikelihood penalty to the loss has demonstrated success in reducing unwanted degenerative traits in models  \citep{welleck2019neural,li-etal-2020-dont}. For example, to prevent \textit{repetitions}, an unwanted token set of previous contexts, $k \in \mathcal{C}_{recent}$, is disfavored by adding the following term to $(1-p_i^{(k)})$ to the loss (such as Equation \ref{eq:celoss}) which penalizes the model for assigning high probabilities to recent tokens. Motivated by SFT ascribing high probabilities to rejected tokens (Figure \ref{fig:sft}) and the effectiveness of appending penalizing unwanted traits, we design a monolithic preference alignment method that dynamically penalizes the disfavored response for each query without the need for crafting sets of rejected tokens.

\section{Odds Ratio Preference Optimization}\label{sec:method}
We introduce a novel preference alignment algorithm, Odds Ratio Preference Optimization (\texttt{ORPO}), which incorporates an odds ratio-based penalty to the conventional negative log-likelihood (NLL) loss for differentiating the generation styles between favored and disfavored responses.

\subsection{Preliminaries}
Given an input sequence $x$, the average log-likelihood of generating the output sequence $y$, of length $m$ tokens, is computed as Equation \ref{eq:likeli}. The odds of generating the output sequence $y$ given an input sequence $x$ is defined in Equation \ref{eq:odds}:
\begin{equation}
    \log P_\theta(y|x) = \frac{1}{m} \sum_{t=1}^m \log P_\theta(y_t|x, y_{<t})\label{eq:likeli}
\end{equation}
\begin{equation}
    \textbf{odds}_\theta(y|x) = \frac{P_\theta(y|x)}{1 - P_\theta(y|x)}\label{eq:odds}
\end{equation}
Intuitively, $\textbf{odds}_\theta(y|x) = k$ implies that it is $k$ times more likely for the model $\theta$ to generate the output sequence $y$ than not generating it. Thus, the odds ratio of the chosen response $y_w$ over the rejected response $y_l$, $\textbf{OR}_\theta(y_w, y_l)$, indicates how much more likely it is for the model $\theta$ to generate $y_w$ than $y_l$ given input $x$, defined in Equation \ref{eq:or}.
\begin{equation}
    \textbf{OR}_\theta(y_w, y_l) = \frac{\textbf{odds}_\theta(y_w|x)}{\textbf{odds}_\theta(y_l|x)}\label{eq:or}
\end{equation}

\subsection{Objective Function of \texttt{ORPO}}

The objective function of \texttt{ORPO} in Equation~\ref{eq:main} consists of two components: 1) supervised fine-tuning (SFT) loss ($\mathcal{L}_{SFT}$); 2) relative ratio loss ($\mathcal{L}_{OR}$).
\begin{equation}
    \mathcal{L}_{ORPO} = \mathbb{E}_{(x, y_w, y_l)}\left[ \mathcal{L}_{SFT} + \lambda \cdot \mathcal{L}_{OR} \right]
    \label{eq:main}
\end{equation}
$\mathcal{L}_{SFT}$ follows the conventional causal language modeling negative log-likelihood (NLL) loss function to maximize the likelihood of generating the reference tokens as previously discussed in Section~\ref{sec:role}.
$\mathcal{L}_{OR}$ in Equation \ref{eq:ratio} maximizes the odds ratio between the likelihood of generating the disfavored response $y_w$ and the disfavored response $y_l$. We wrap the log odds ratio with the log sigmoid function so that $\mathcal{L}_{OR}$ could be minimized by increasing the log odds ratio between $y_w$ and $y_l$.
\begin{equation}
    \mathcal{L}_{OR} = -\log \sigma \left( \log \frac{\textbf{odds}_\theta(y_w|x)}{\textbf{odds}_\theta(y_l|x)} \right) \label{eq:ratio} 
\end{equation}
Together, $\mathcal{L}_{SFT}$ and $\mathcal{L}_{OR}$ weighted with $\lambda$ tailor the pre-trained language model to adapt to the specific subset of the desired domain and disfavor generations in the rejected response sets.

\subsection{Gradient of \texttt{ORPO}}\label{subsec:grad}

The gradient of $\mathcal{L}_{Ratio}$ further justifies using the odds ratio loss. It comprises two terms: one that penalizes the wrong predictions and one that contrasts between chosen and rejected responses, denoted in Equation \ref{eq:simp}\footnote{The full derivation for $\nabla_\theta \mathcal{L}_{OR}$ is in Appendix \ref{ap:derivative}.} for $d=(x, y_l, y_w)\sim D$.
\begin{equation}
    \nabla_\theta \mathcal{L}_{OR} = \delta(d) \cdot h(d)\label{eq:simp}
\end{equation}
\begin{align}
    \delta(d) &= \left[ 1 + \frac{\textbf{odds}_\theta P(y_w|x)}{\textbf{odds}_\theta P(y_l|x)} \right]^{-1}\label{eq:delta} \\
    h(d) &= \frac{\nabla_\theta \log P_\theta(y_w|x)}{1 - P_\theta(y_w|x)} - \frac{\nabla_\theta \log P_\theta(y_l|x)}{1 - P_\theta(y_l|x)}\label{eq:grad}
\end{align}
When the odds of the favored responses are relatively higher than the disfavored responses, $\delta(d)$ in Equation \ref{eq:delta} will converge to 0. This indicates that the $\delta(d)$ will play the role of a penalty term, accelerating the parameter updates if the model is more likely to generate the rejected responses. 

Meanwhile, $h(d)$ in Equation \ref{eq:grad} implies a weighted contrast of the two gradients from the chosen and rejected responses. Specifically, $1-P(y|x)$ in the denominators amplifies the gradients when the corresponding side of the likelihood $P(y|x)$ is low. For the chosen responses, this accelerates the model's adaptation toward the distribution of chosen responses as the likelihood increases.

\section{Experimental Settings}

\subsection{Training Configurations}
\paragraph{Models} We train a series of OPT models \citep{zhang2022opt} scaling from 125M to 1.3B parameters comparing supervised fine-tuning (SFT), proximal policy optimization (PPO), direct policy optimization (DPO), and compare these to our \texttt{ORPO}. PPO and DPO models were fine-tuned with TRL library \citep{vonwerra2022trl} on top of SFT models trained for a single epoch on the chosen responses following \citet{rafailov2023direct} and \citet{tunstall2023zephyr}. We notate this by prepending "+" to each algorithm (e.g., +DPO). Additionally, we train Phi-2 (2.7B) \citep{Javaheripi_Bubeck_2023}, a pre-trained language model with promising downstream performance \citep{open-llm-leaderboard}, as well as Llama-2 (7B) \citep{touvron2023llama} and Mistral (7B) \citep{jiang2023mistral}.
Further training details for each method are in Appendix \ref{ap:detail}.

\paragraph{Datasets} We test each training configuration and model on two datasets: 1) Anthropic's HH-RLHF \citep{bai2022training}, 2) Binarized UltraFeedback \citep{tunstall2023zephyr}. We filtered out instances where $y_w=y_l$ or where $y_w=\emptyset$ or where $y_l=\emptyset$.

\paragraph{Reward Models} We train OPT-350M and OPT-1.3B on each dataset for a single epoch for reward modeling with the objective function in Equation \ref{eq:rm} \citep{ziegler2020finetuning}. The OPT-350M reward model was used for PPO, and OPT-1.3B reward model was used to assess the generations of fine-tuned models. We refer to these reward models as RM-350M and RM-1.3B in Section \ref{sec:result}.
\begin{equation}
    -\mathbb{E}_{(x, y_l, y_w)} \left[ \log \sigma \left(r(x, y_w) - r(x, y_l) \right) \right]
    \label{eq:rm}
\end{equation}

\subsection{Leaderboard Evaluation} In Section~\ref{subsec:instruct},  we evaluate the models using the $\text{AlpacaEval}_{1.0}$ and $\text{AlpacaEval}_{2.0}$ \citep{alpaca_eval} benchmarks, comparing \texttt{ORPO} to other instruction-tuned models reported in the official leaderboard,\footnote{\url{https://tatsu-lab.github.io/alpaca_eval/}} including Llama-2 Chat (7B) and (13B) \citep{touvron2023llama}, and Zephyr $\alpha$ and $\beta$ \citep{almazrouei2023falcon}. Similarly, in Section \ref{subsec:multi}, we evaluate the models with MT-Bench \citep{zheng_judging_2023} and report the results and the scores of the same models reported in the official leaderboard.\footnote{\url{https://huggingface.co/spaces/lmsys/chatbot-arena-leaderboard}}
Using GPT-4 \citep{achiam2023gpt} as an evaluator in $\text{AlpacaEval}_{1.0}$, we assess if the trained model can be preferred over the responses generated from \texttt{text-davinci-003}. For $\text{AlpacaEval}_{2.0}$, we used GPT-4-turbo\footnote{\url{https://platform.openai.com/docs/models/gpt-4-and-gpt-4-turbo}} as an evaluator following the default setting. We assess if the generated responses are favored over those generated from GPT-4. Finally, using GPT-4 as an evaluator in MT-Bench, we check if the models can follow the instructions with hard answers in a multi-turn conversation.

\section{Results and Analysis}\label{sec:result}

First, we assess the general instruction-following abilities of the models by comparing the preference alignment algorithms in Sections \ref{subsec:instruct} and \ref{subsec:multi}. Second, we measure the win rate of OPT models trained with \texttt{ORPO} against other alignment methods training OPT 1.3B as a reward model in Section \ref{subsec:rmwin}. Then, we measure the lexical diversity of the models trained with \texttt{ORPO} and DPO in Section \ref{subsec:div}.

\subsection{Single-turn Instruction Following}\label{subsec:instruct}

\begin{table*}[hbt!]
\setlength{\tabcolsep}{12pt}
\centering
\resizebox{350pt}{!}{
\begin{tabular}{l|c|c|c}
\hline \hline
\textbf{Model Name}       & \textbf{Size} &  $\textbf{AlpacaEval}_{\textbf{1.0}}$    &$\textbf{AlpacaEval}_{\textbf{2.0}}$   \\ \hline

Phi-2 + SFT                 &  2.7B  & 48.37\% (1.77)          & 0.11\% (0.06) \\
Phi-2 + SFT + DPO           &  2.7B  &   50.63\% (1.77)           & 0.78\% (0.22) \\
Phi-2 + \texttt{ORPO} \textit{(Ours)}    &  2.7B  &  \textbf{71.80\% (1.59)}  & \textbf{6.35\% (0.74)} \\  \hline  
Llama-2 Chat *              &  7B    &  71.34\% (1.59)          & 4.96\% (0.67) \\  
Llama-2 Chat *              &  13B   &  81.09\% (1.38)          & 7.70\% (0.83) \\ 
Llama-2 + \texttt{ORPO} \textit{(Ours)}     &  7B    &  \textbf{81.26\% (1.37)}          & \textbf{9.44\% (0.85)}  \\ \hline
Zephyr ($\alpha$) *         &  7B    &  85.76\% (1.23)          & 8.35\% (0.87) \\  
Zephyr ($\beta$) *          &  7B    &  90.60\% (1.03)          & 10.99\% (0.96) \\ 
Mistral-\texttt{ORPO}-$\alpha$ \textit{(Ours)}     &  7B    &   87.92\% (1.14)   & 11.33\% (0.97)  \\ 
Mistral-\texttt{ORPO}-$\beta$ \textit{(Ours)}     &  7B    &   \textbf{91.41\% (1.15)}   & \textbf{12.20\% (0.98)}  \\ \hline \hline

\end{tabular}%
}
\caption{Table of instruction-following abilities of each checkpoint measured through AlpacaEval. While clearly showing the improvements in instruction-following abilities after training with \texttt{ORPO}, it is notable that \texttt{ORPO} models exceed RLHF or DPO models of Llama-2 and Mistral (* indicates the results from the official leaderboard.)}
\label{tab:result}
\end{table*}

\paragraph{Phi-2 (2.7B)} \texttt{ORPO} improved pre-trained Phi-2 to exceed the performance of the Llama-2 Chat instruction-following language model by \textit{only using UltraFeedback} as the instruction-tuning dataset, as shown in Table \ref{tab:result}. $\lambda$ of 0.25 was applied for Phi-2, resulting in 71.80\% and 6.35\% in AlpacaEval.

\paragraph{Llama-2 (7B)} Notably, UltraFeedback and \texttt{ORPO} with $\lambda$ of 0.2 on Llama-2 (7B) resulted in higher AlpacaEval scores than the chat versions of both 7B and 13B scale trained with RLHF, eventually showing 81.26\% and 9.44\% in both AlpacaEvals.

In contrast, in our controlled experimental setting of conducting one epoch of SFT and three epochs of DPO following \citet{tunstall2023zephyr} and \citet{rafailov2023direct}, Llama-2 + SFT and Llama-2 + SFT + DPO yielded models with outputs that could not be evaluated. This supports the efficacy of \texttt{ORPO}, in which the model can rapidly learn the desired domain and the preference with limited data. This aligns with the $h(d)$ examination in the gradient of our method studied in Section \ref{subsec:grad}.

\paragraph{Mistral-\texttt{ORPO}-$\alpha$ (7B)} Furthermore, fine-tuning Mistral (7B) with single-turn conversation dataset, UltraFeedback, and \texttt{ORPO} with $\lambda$ of 0.1 outperforms Zephyr series, which are the Mistral (7B) models fine-tuned with SFT on 20K UltraChat \citep{ding2023enhancing} and DPO on the full UltraFeedback. As shown in Table \ref{tab:result}, Mistral-\texttt{ORPO}-$\alpha$ (7B) achieves 87.92\% and 11.33\%, which exceeds Zephyr $\alpha$ by 1.98\% and Zephyr $\beta$ by 0.34\% in $\text{AlpacaEval}_{2.0}$. The sample responses and corresponding references from GPT-4 can be found in Appendix \ref{ap:generation}.

\paragraph{Mistral-\texttt{ORPO}-$\beta$ (7B)} Using the same configuration of Mistral-\texttt{ORPO}-$\alpha$ (7B), we additionally compare fine-tuning Mistral on the cleaned version of the UltraFeedback\footnote{\url{https://huggingface.co/datasets/argilla/ultrafeedback-binarized-preferences-cleaned}} to demonstrate the effect of the data quality \citep{notus2023}. While the actual sizes of datasets are similar, \texttt{ORPO} gains further advantages from the dataset quality by scoring over 91\% and 12\% on AlpacaEval, as shown in Table \ref{tab:result}. Further instruction-following evaluation on two Mistral-based models with IFEval \citep{zhou2023instructionfollowing} is reported in the Appendix \ref{ap:ifeval}.

\subsection{Multi-turn Instruction Following}\label{subsec:multi}

With our best model, Mistral-\texttt{ORPO}-$\alpha$ (7B) and Mistral-\texttt{ORPO}-$\beta$ (7B), we also assess the multi-turn instruction-following skills with deterministic answers (e.g., math) through MT-Bench. 
\begin{figure}[t]
    \centering
    \includegraphics[width=\columnwidth]{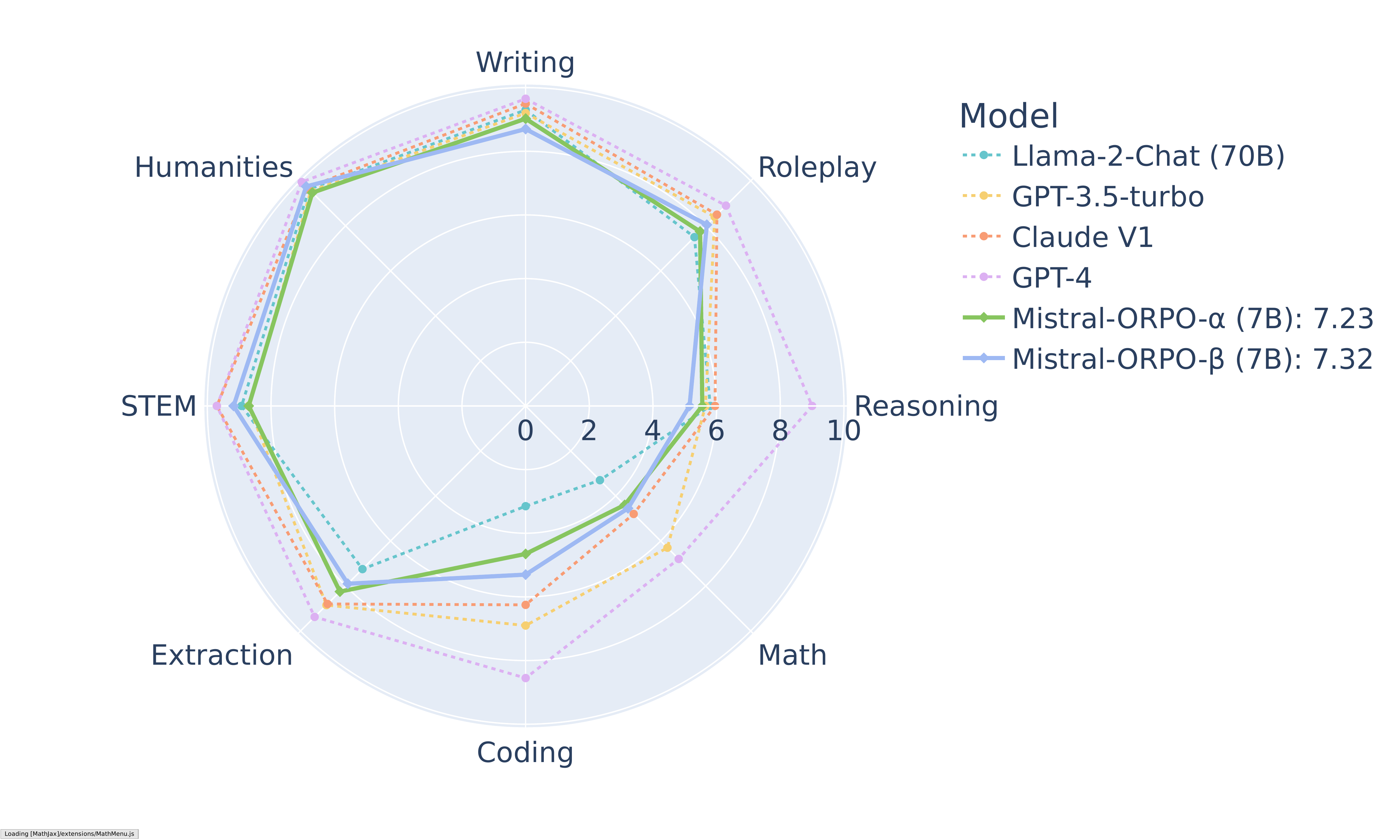}
    \caption{MT-Bench result of Mistral-\texttt{ORPO}-$\alpha$ (7B) and Mistral-\texttt{ORPO}-$\beta$ (7B) by the category. Further comparison can be found in the Appendix \ref{ap:mtbench}.}
    \label{fig:mtbench-main}
\end{figure}

As shown in Figure \ref{fig:mtbench-main}, \texttt{ORPO}-Mistral (7B) series achieve comparable results to either larger or the proprietary models, including Llama-2-Chat (70B) and Claude. Eventually, Mistral-\texttt{ORPO}-$\alpha$ (7B) and Mistral-\texttt{ORPO}-$\beta$ (7B) scored 7.23 and 7.32 in MT-Bench without being exposed to the multi-turn conversation dataset during training.

\begin{figure*}[hbt!]
    \centering
    \includegraphics[width=\textwidth]{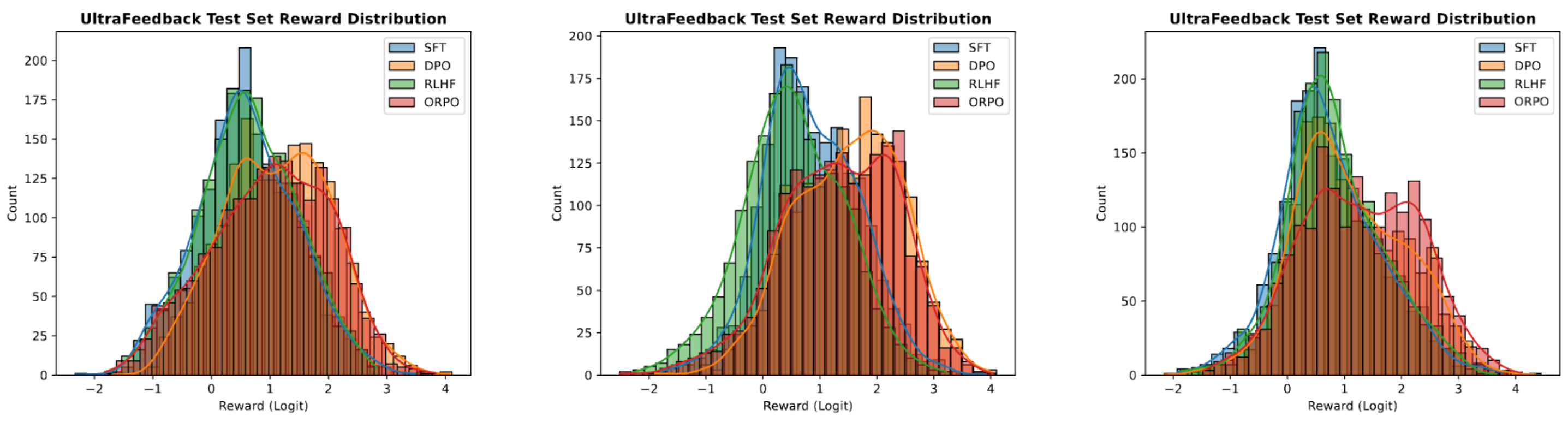}
    \caption{Reward distribution comparison between OPT-125M (left), OPT-350M (middle), and OPT-1.3B (right) trained with SFT (blue), RLHF (green), DPO (orange), and \texttt{ORPO} (red) on the test set of UltraFeedback using the RM-1.3B. While the rewards of the trained models are roughly normal and preference optimization algorithms (RLHF, DPO, and \texttt{ORPO}) tend to move the reward distribution in the positive direction, \texttt{ORPO} is on par or better than RLHF and DPO in increasing the expected reward. The same plot for the HH-RLHF dataset is in Appendix \ref{ap:hh}.}
    \label{fig:foobar}
\end{figure*}

\subsection{Reward Model Win Rate}\label{subsec:rmwin}

We assess the win rate of \texttt{ORPO} over other preference alignment methods, including supervised fine-tuning (SFT), PPO, and DPO, using RM-1.3B to understand the effectiveness and scalability of \texttt{ORPO} in Tables \ref{tab:rm_win_hh} and \ref{tab:rm_win_uf}. Additionally, we visually verify that \texttt{ORPO} can effectively enhance the expected reward compared to SFT in Figure \ref{fig:foobar}.

\paragraph{HH-RLHF} In Table \ref{tab:rm_win_hh}, \texttt{ORPO} outperforms SFT and PPO across all model scales. The highest win rate against SFT and PPO across the size of the model was 78.0\% and 79.4\%, respectively. Meanwhile, the win rate over DPO was correlated to the model's size, with the largest model having the highest win rate: 70.9\%.

\begin{table}[hbt!]
\resizebox{\columnwidth}{!}{%
\begin{tabular}{c|ccc}
\hline \hline
\textbf{\texttt{ORPO} vs}  & \textbf{SFT}    & \textbf{+DPO}      & \textbf{+PPO}      \\ \hline
\textbf{OPT-125M} & 84.0 (0.62)     & 41.7 (0.77)       & 66.1 (0.26) \\
\textbf{OPT-350M} & 82.7 (0.56)     & 49.4 (0.54)       & 79.4 (0.29)  \\
\textbf{OPT-1.3B} & 78.0 (0.16)     & 70.9 (0.52)       & 65.9 (0.33)    \\ \hline \hline
\end{tabular}%
}
\caption{Average win rate (\%) and its standard deviation of \texttt{ORPO} and standard deviation over other methods on \textbf{HH-RLHF} dataset for three rounds. Sampling decoding with a temperature of 1.0 was used on the test set.}
\label{tab:rm_win_hh}
\end{table}

\paragraph{UltraFeedback} The win rate in UltraFeedback followed similar trends to what was reported in HH-RLHF, as shown in Table \ref{tab:rm_win_uf}. \texttt{ORPO} was preferred over SFT and PPO for maximum 80.5\% and 85.8\%, respectively. While consistently preferring \texttt{ORPO} over SFT and PPO, the win rate over DPO gradually increases as the size of the model increases. The scale-wise trend exceeding DPO will be further shown through 2.7B models in Section \ref{subsec:instruct}.
\begin{table}[hbt!]
\resizebox{\columnwidth}{!}{%
\begin{tabular}{c|ccc}
\hline \hline
\textbf{\texttt{ORPO} vs}  & \textbf{SFT}    & \textbf{+DPO}      & \textbf{+PPO}      \\ \hline
\textbf{OPT-125M} & 73.2 (0.12)     & 48.8 (0.29)       & 71.4 (0.28) \\
\textbf{OPT-350M} & 80.5 (0.54)     & 50.5 (0.17)       & 85.8 (0.62)  \\
\textbf{OPT-1.3B} & 69.4 (0.57)     & 57.8 (0.73)       & 65.7 (1.07)  \\ \hline \hline
\end{tabular}%
}
\caption{Average win rate (\%) and its standard deviation of \texttt{ORPO} and standard deviation over other methods on \textbf{UltraFeedback} dataset for three rounds. Sampling decoding with a temperature of 1.0 was used.}
\label{tab:rm_win_uf}
\end{table}

\paragraph{Overall Reward Distribution} In addition to the win rate, we compare the reward distribution of the responses generated with respect to the test set of the UltraFeedback dataset in Figure \ref{fig:foobar} and HH-RLHF dataset in Appendix \ref{ap:hh}. Regarding the SFT reward distribution as a default, PPO, DPO, and \texttt{ORPO} shift it in both datasets. However, the magnitude of reward shifts for each algorithm differs.

In Figure \ref{fig:foobar}, RLHF (i.e., SFT + PPO) has some abnormal properties of the distribution with a low expected reward. We attribute this to empirical evidence of the instability and reward mismatch problem of RLHF \citep{rafailov2023direct, gao2022scaling, shen-etal-2023-loose} as the RLHF models were trained with RM-350M and assessed with RM-1.3B. Meanwhile, it is notable that the \texttt{ORPO} distribution (red) is mainly located on the very right side of each subplot, indicating higher expected rewards. Recalling the intent of preference alignment methods, the distributions in Figure \ref{fig:foobar} indicate that \texttt{ORPO} tends to fulfill the aim of preference alignment for all model sizes.

\subsection{Lexical Diversity}\label{subsec:div}

The lexical diversity of the preference-aligned language models was studied in previous works \citep{kirk2024understanding}. We expand the concept of per-input and across-input diversity introduced in \citet{kirk2024understanding} by using Gemini-Pro \citep{team2023gemini} as an embedding model, which is suitable for assessing the diversity of instruction-following language models by encoding a maximum of 2048 tokens. The diversity metric with the given set of sampled responses is defined as Equation \ref{eq:distance}.
\begin{equation}
    \mathcal{O}_\theta^i := \{y_j \sim \theta(y|x_i) | j = 1, 2, ..., K \}\label{eq:set}
\end{equation}
\begin{equation}
    D(\mathcal{O}^i_\theta) = \frac{1}{2} \cdot \frac{\sum_{i=1}^{N-1} \sum_{j=i+1}^{N} \cos(h_i, h_j)}{N \cdot (N-1)}\label{eq:distance}
\end{equation}

where $\cos(h_i, h_j)$ refers to the cosine similarity between the embedding $h_i$ and $h_j$. 5 different responses are sampled with a temperature of 1.0 to 160 queries in AlpacaEval (i.e., $K=5, N=160$) using Phi-2 and Llama-2 trained with \texttt{ORPO} and DPO. We report the results in Table \ref{tab:diversity}.

\paragraph{Per Input Diversity (PID)} We average the input-wise average cosine similarity between the generated samples with Equation \ref{eq:pid} to assess the per-input diversity. In Table \ref{tab:diversity}, \texttt{ORPO} models have the highest average cosine similarity in the first column for both models, which implies the lowest diversity per input. This indicates that \texttt{ORPO} generally assigns high probabilities to the desired tokens, while DPO has a relatively smoother logit distribution.

\begin{equation}
    \text{PID}_D(\theta) = \frac{1}{N}\sum_{i=1}^N D(\mathcal{O}^i_\theta)\label{eq:pid}
\end{equation}

\paragraph{Across Input Diversity (AID)} Using 8 samples generated per input, we sample the first item for each input and examine their inter cosine similarity with Equation \ref{eq:aid} for across-input diversity. Unlike per-input diversity, it is noteworthy that Phi-2 (\texttt{ORPO}) has lower average cosine similarity in the second row of Table \ref{tab:diversity}. We can infer that \texttt{ORPO} triggers the model to generate more instruction-specific responses than DPO.

\begin{equation}
    \text{AID}_D(\theta) = D \left( \bigcup\limits_{i=1}^{N}\mathcal{O}^i, _{\theta, j=1} \right)\label{eq:aid}
\end{equation}

\begin{table}[hbt!]
\centering
\resizebox{\columnwidth}{!}{%
\begin{tabular}{l|cc}
\hline \hline
\multicolumn{1}{c|}{}       & \textbf{Per Input}$\downarrow$ & \textbf{Across Input}$\downarrow$ \\ \hline
Phi-2 + SFT + DPO    & \textbf{0.8012}             & 0.6019                \\
Phi-2 + ORPO         & 0.8909             & \textbf{0.5173}                \\ \hline 
Llama-2 + SFT + DPO  & \textbf{0.8889}             & 0.5658               \\
Llama-2 + ORPO       & 0.9008             & \textbf{0.5091}                \\ \hline \hline
\end{tabular}%
}
\caption{Lexical diversity of Phi-2 and Llama-2 fine-tuned with DPO and \texttt{ORPO}. Lower cosine similarity is equivalent to higher diversity. The highest value in each column within the same model family is bolded.}
\label{tab:diversity}
\end{table}

\section{Discussion}

In this section, we expound on the theoretical and computational details of \texttt{ORPO}. The theoretical analysis of $\texttt{ORPO}$ is studied in Section \ref{subsec:why}, which will be supported with the empirical analysis in Section \ref{subsec:odds_trend}. Then, we compare the computational load of DPO and \texttt{ORPO} in Section \ref{subsec:compute}.

\subsection{Comparison to Probability Ratio}\label{subsec:why}

The rationale for selecting the odds ratio instead of the probability ratio lies in its stability. The probability ratio for generating the favored response $y_w$ over the disfavored response $y_l$ given an input sequence $x$ can be defined as Equation \ref{eq:pr}. 
\begin{equation}
    \textbf{PR}_\theta(y_w, y_l) = \frac{P_\theta(y_w|x)}{P_\theta(y_l|x)}\label{eq:pr}
\end{equation}

While this formulation has been used in previous preference alignment methods that precede SFT \citep{rafailov2023direct, azar2023general}, the odds ratio is a better choice in the setting where the preference alignment is incorporated in SFT as the odds ratio is more sensitive to the model's preference understanding. In other words, the probability ratio leads to more extreme discrimination of the disfavored responses than the odds ratio.

\begin{figure}[hbt!]
    \centering
    \includegraphics[width=\columnwidth]{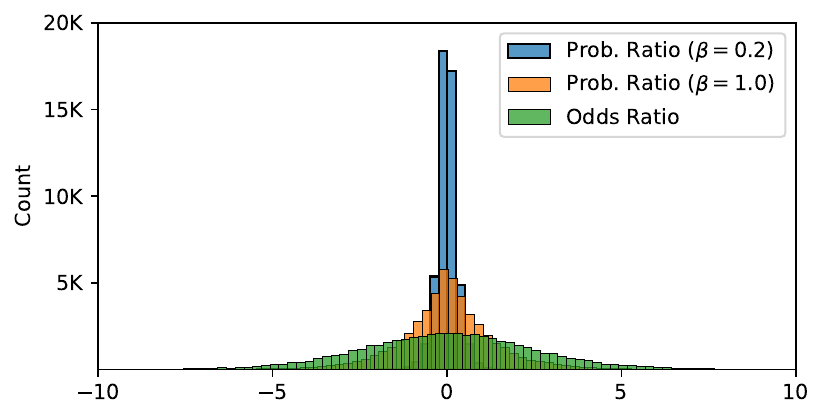}
    \caption{Sampled distribution of $\log \textbf{PR}(X_2|X_1)$ and $\log \textbf{OR}(X_2|X_1)$. $\log \textbf{OR}(X_2|X_1)$ has a wider range given the same input probability pairs $(X_1, X_2)$.}
    \label{fig:compare}
\end{figure}

We visualize this through the sample distributions of the log probability ratio $\log \textbf{PR}(X_2|X_1)$ and log odds ratio $\log \textbf{OR}(X_2|X_1)$. We sample 50,000 samples each with Equation \ref{eq:unif} and plot the log probability ratio and log odds ratio in Figure \ref{fig:compare}. We multiply $\beta$ for the probability ratio as it is practiced in the probability ratio-based methods and report the cases where $\beta=0.2$ and $\beta=1.0$.
\begin{gather}
    X_1, X_2 \sim \text{Unif}(0, 1)\label{eq:unif} \\
    Y \sim \beta \left( \log X_1 - \log X_2 \right)\label{eq:sample_prob} \\
    Y \sim \log \frac{X_1}{1 - X_1} - \log \frac{X_2}{1 - X_2}\label{eq:sample_odds}
\end{gather}

Recalling that the log sigmoid function is applied to the log probability ratio and log odds ratio, each ratio's scale determines the expected margin between the likelihood of the favored and disfavored styles when the loss is minimized. In that sense, the contrast should be relatively extreme to minimize the log sigmoid loss when $\textbf{PR}(X_2|X_1)$ is inputted instead of $\textbf{OR}(X_2|X_1)$ to the log sigmoid function, regarding the sharp distribution of $\log \textbf{PR}(X_2|X_1)$ in Figure \ref{fig:compare}. This results in overly suppressing the logits for the tokens in the disfavored responses in the setting where SFT and preference alignment are incorporated, as the model is not adapted to the domain. We empirically support this analysis through the ablation study in Appendix \ref{ap:rpor}. Therefore, the odds ratio is a better choice when the preference alignment is done with SFT  due to the mild discrimination of disfavored responses and the prioritizing of the favored responses to be generated.

Throughout fine-tuning, minimizing the log sigmoid loss leads to either $\textbf{PR}(X_2|X_1)$ or $\textbf{OR}(X_2|X_1)$ to be larger. This is equivalent to the rejected responses' token-wise likelihood, which will generally get smaller. In this context, it is essential to avoid an overly extreme contrast. This precaution is especially important given the sharp distribution of $\log \textbf{PR}(X_2|X_1)$ depicted in Figure \ref{fig:compare}. The excessive margin could lead to the unwarranted suppression of logits for tokens in disfavored responses within the incorporated setting, potentially resulting in issues of degeneration.

\subsection{Minimizing $\mathcal{L}_{OR}$}\label{subsec:odds_trend}
\begin{figure}[hbt!]
    \centering
    \includegraphics[width=\columnwidth]{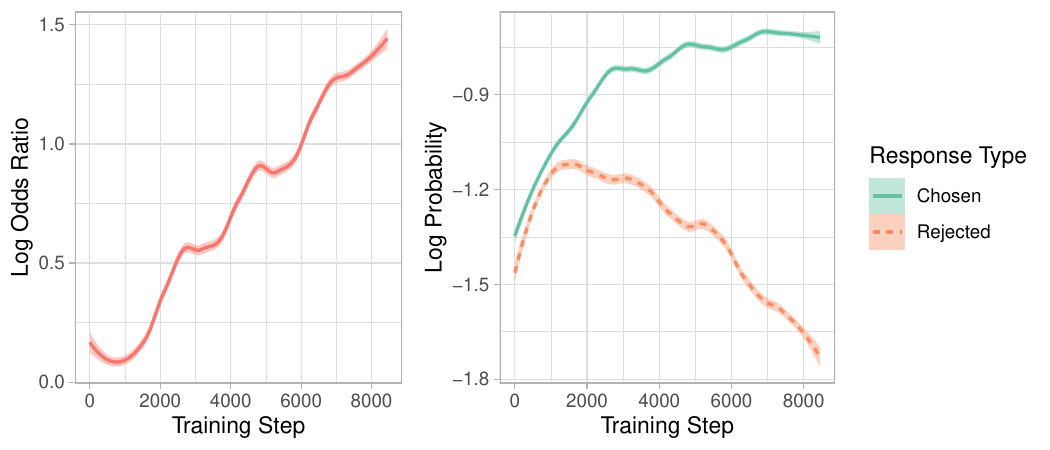}
    \caption{Average log-likelihood for chosen and rejected responses and log odds ratio per batch. The odds consistently increase during training with \texttt{ORPO}.}
    \label{fig:log_odds}
\end{figure}

We demonstrate that models trained with \texttt{ORPO} learned to reflect the preference throughout the training process. We monitored the log probabilities of the chosen and rejected responses and the log odds ratio with $\lambda = 1.0$. With the same dataset and model as Figure \ref{fig:sft}, Figure \ref{fig:log_odds} shows that the log probability of rejected responses is diminishing while that of chosen responses is on par with Figure \ref{fig:sft} as the log odds ratio increases. This indicates that \texttt{ORPO} is successfully preserving the domain adaptation role of SFT while the penalty term $L_{OR}$ induces the model to lower the likelihood of unwanted generations. We discuss the effect of $\lambda$ in Equation \ref{eq:main} in Appendix \ref{ap:ablation}, studying the proclivity of the log probability margin between the favored and disfavored responses with respect to $\lambda$.

\subsection{Computational Efficiency}\label{subsec:compute}

As depicted in Figure \ref{fig:main}, \texttt{ORPO} does not require a reference model, unlike RLHF and DPO. In that sense, \texttt{ORPO} is computationally more efficient than RLHF and DPO in two perspectives: 1) memory allocation and 2) fewer FLOPs per batch.

The reference model ($\pi_{SFT}$) in the context of RLHF and DPO denotes the model trained with supervised fine-tuning (SFT), which will be the baseline model for updating the parameters with RLHF or DPO \citep{ziegler2020finetuning,rafailov2023direct}. Thus, two $\pi_{SFT}$s, a frozen reference model and the model undergoing tuning, are required during training. Furthermore, in theory, two forward passes should be calculated for each model to acquire the logits for the chosen and rejected responses. In other words, four forward passes happen in total for a single batch. 
On the other hand, a reference model is not required in \texttt{ORPO} as $\pi_{SFT}$ is directly updated. This leads to half the number of forward passes required for each batch during training.

\section{Conclusion}

In this paper, we introduced a reference-free monolithic preference alignment method, odds ratio preference optimization (\texttt{ORPO}), by revisiting and understanding the value of the supervised fine-tuning (SFT) phase in the context of preference alignment. \texttt{ORPO} was consistently preferred by the fine-tuned reward model against SFT and RLHF across the scale, and the win rate against DPO increased as the size of the model increased. Furthermore, we validate the scalability of \texttt{ORPO} with 2.7B and 7B pre-trained language models by exceeding the larger state-of-the-art instruction-following language models in AlpacaEval. Specifically, Mistral-\texttt{ORPO}-$\alpha$ and Mistral-\texttt{ORPO}-$\beta$ achieved 11.33\% and 12.20\% in $\text{AlpacaEval}_{2.0}$, 7.23 and 7.32 in MT-Bench, thereby underscoring the efficiency and effectiveness of \texttt{ORPO}. We release fine-tuning code and model checkpoints for Mistral-\texttt{ORPO}-$\alpha$ and Mistral-\texttt{ORPO}-$\beta$ to aid reproducibility.

\section*{Limitations}

While conducting a comprehensive analysis of the diverse preference alignment methods, including DPO and RLHF, we did not incorporate a more comprehensive range of preference alignment algorithms. We leave the broader range of comparison against other methods as future work, along with scaling our method to over 7B models. In addition, we will expand the fine-tuning datasets into diverse domains and qualities, thereby verifying the generalizability of our method in various NLP downstream tasks. Finally, we would like to study the internal impact of our method on the pre-trained language model, expanding the understanding of preference alignment procedure to not only the supervised fine-tuning stage but also consecutive preference alignment algorithms.

\bibliography{anthology,custom}
\bibliographystyle{acl_natbib}

\setlength{\jot}{10pt}

\onecolumn
\appendix

\section{Derivation of $\nabla_\theta \mathcal{L}_{OR}$ with Odds Ratio}
\label{ap:derivative}

Suppose that $g(x, y_l, y_w) = \frac{\textbf{odds}_\theta P(y_w|x)}{\textbf{odds}_\theta P(y_l|x)}$

\begin{align}
    \nabla_\theta \mathcal{L}_{OR} &= \nabla_\theta \log \sigma \left( \log \frac{\textbf{odds}_\theta P(y_w|x)}{\textbf{odds}_\theta P(y_l|x)}\right)\\
    &=\frac{\sigma'\left(\log g(x, y_l, y_w) \right)}{\sigma\left(\log g(x, y_l, y_w) \right)} \\
    &=\sigma\left(-\log g(x, y_l, y_w)\right) \cdot \nabla_\theta \log g(x, y_l, y_w) \\
    &= \frac{\sigma\left(-\log g(x, y_l, y_w)\right)}{g(x, y_l, y_w)} \cdot \nabla_\theta g(x, y_l, y_w) \\
    &= \sigma\left(-\log g(x, y_l, y_w)\right) \cdot \nabla_\theta \log g(x, y_l, y_w) \\
    &= \left(1 + \frac{\textbf{odds}_\theta P(y_w|x)}{\textbf{odds}_\theta P(y_l|x)} \right)^{-1} \cdot \nabla_\theta \log \frac{\textbf{odds}_\theta P(y_w|x)}{\textbf{odds}_\theta P(y_l|x)}\label{eq:mid_result}
\end{align}

In Equation \ref{eq:mid_result}, the remaining derivative can be further simplified by replacing $1 - P_\theta(y|x)$ terms where $P(y|x) = \sqrt[N]{\prod_t^N P_\theta(y_t|x, y_{<t}}$ in $\textbf{odds}_\theta (y|x)$ as follows.

\begin{align}
    \nabla_\theta \log \left(1 - P_\theta(y|x)\right) &= \frac{\nabla_\theta \left(1 - P_\theta(y|x)\right)}{1 - P_\theta(y|x)} \\
    &= \frac{-\nabla_\theta P_\theta(y|x)}{1 - P_\theta(y|x)} \\
    &= -\frac{P_\theta(y|x)}{1 - P_\theta(y|x)} \cdot \nabla_\theta \log P_\theta(y|x) \\
    &= \textbf{odds}_\theta(y|x) \cdot \nabla_\theta \log P_\theta(y|x)\label{eq:one_minus}\\
    \nabla_\theta \log \frac{\textbf{odds}_\theta P(y_w|x)}{\textbf{odds}_\theta P(y_l|x)}&= \nabla_\theta \log \frac{P_\theta(y_w|x)}{P_\theta(y_l|x)} - \Bigl( \nabla_\theta \log (1 - P_\theta(y_w|x)) - \nabla_\theta  \log(1 - P_\theta(y_l|x)) \Bigl)\\
    &=\left(1 + \textbf{odds}_\theta P(y_w|x)\right) \nabla_\theta \log P_\theta(y_w|x) - \left(1 + \textbf{odds}_\theta P(y_l|x)\right) \nabla_\theta \log P_\theta(y_l|x)
\end{align}

Therefore, the final form of $\nabla_\theta \mathcal{L}_{OR}$ would be

\begin{align}
    \nabla_\theta \mathcal{L}_{OR} &= \frac{1 + \textbf{odds}_\theta P(y_w|x)}{1 + \frac{\textbf{odds}_\theta P(y_w|x)}{\textbf{odds}_\theta P(y_l|x)}} \cdot \nabla_\theta \log P_\theta(y_w|x) - \frac{1 + \textbf{odds}_\theta P(y_l|x)}{1 + \frac{\textbf{odds}_\theta P(y_w|x)}{\textbf{odds}_\theta P(y_l|x)}} \cdot \nabla_\theta \log P_\theta(y_l|x) \\
    &= \left(1 + \frac{\textbf{odds}_\theta P(y_w|x)}{\textbf{odds}_\theta P(y_l|x)} \right)^{-1} \cdot \left( \frac{\nabla_\theta \log P_\theta(y_w|x)}{1 - P(y_w|x)} - \frac{\nabla_\theta \log P_\theta(y_l|x)}{1 - P(y_l|x)}\right)\label{eq:final_der}
\end{align}
\clearpage

\section{Ablation on Probability Ratio and Odds Ratio}\label{ap:rpor}
\begin{figure*}[hbt!]
    \centering
    \subfigure{\includegraphics[width=0.49\textwidth]{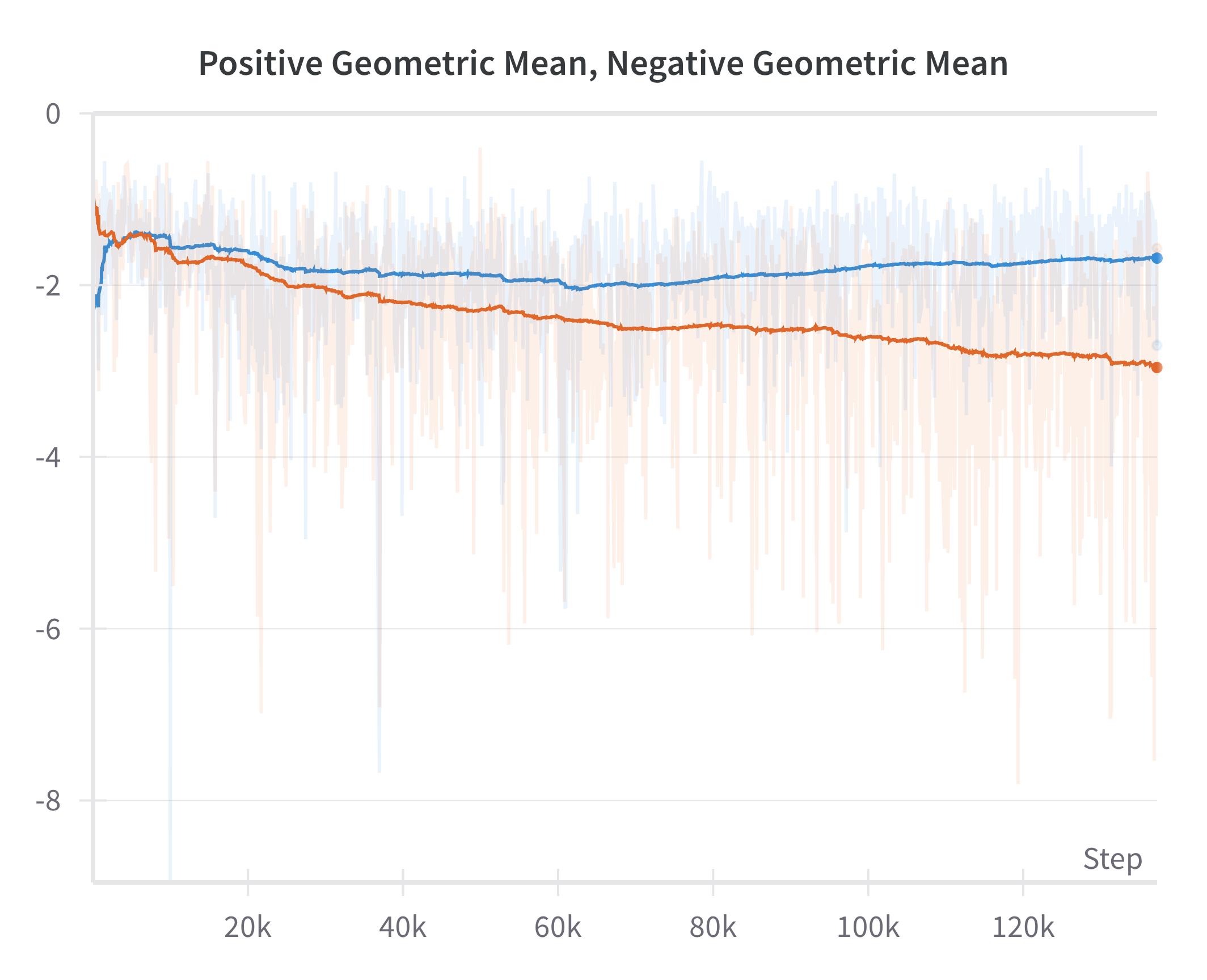}} 
    \subfigure{\includegraphics[width=0.49\textwidth]{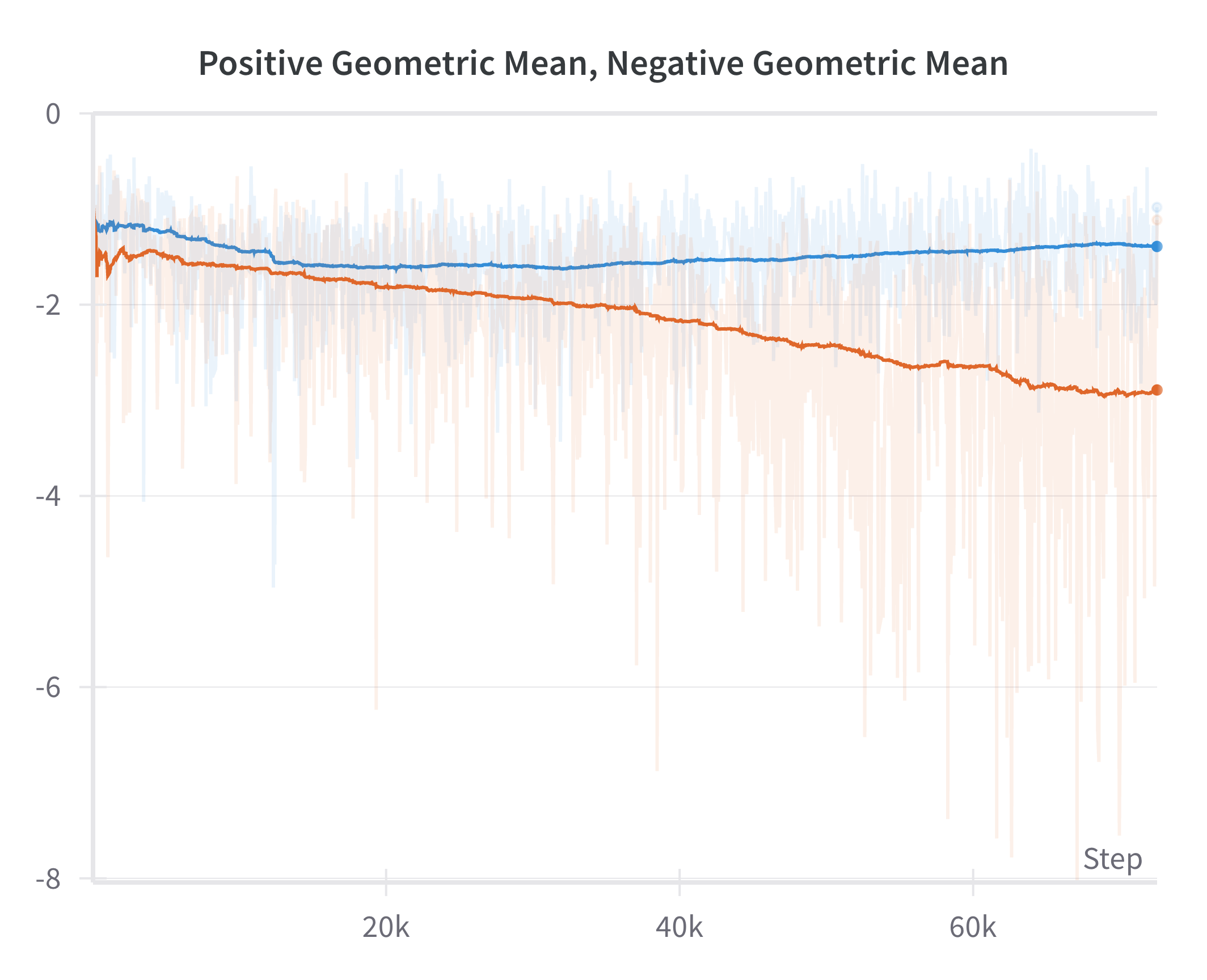}} 
    \caption{The log probability trace when the model is trained with the probability ratio (left) and the odds ratio (right) given the same hyperparameters. The probability ratio leads the rejected responses to have relatively lower log probabilities.}
    \label{fig:ratio_ab}
\end{figure*}

In this section, we continue the discussion in Section \ref{subsec:why} through empirical results comparing the log probabilities of chosen and rejected responses in UltraFeedback when trained with probability ratio and odds ratio. Recalling the sensitivity of each ratio discussed in Section \ref{subsec:why}, it is expected for the probability ratio to lower the log probabilities of the rejected responses with a larger scale than the odds ratio. This is well-shown in Figure \ref{fig:ratio_ab}, which is the log probabilities of each batch while fine-tuning with probability ratio (left) rapidly reaches under -4, while the same phenomenon happens after the over-fitting occurs in the case of odds ratio (right).

\section{Experimental Details}\label{ap:detail}

Flash-Attention 2 \citep{dao2023flashattention2} is applied for all the pre-trained models for computational efficiency. In particular, the OPT series and Phi-2 (2.7B) were trained with DeepSpeed ZeRO 2 \citep{deepspeed}, Llama-2 (7B) and Mistral (7B) were trained with Fully Sharded Data Parallel(FSDP) \citep{zhao2023pytorch}. 7B and 2.7B models were trained with four and two NVIDIA A100, and the rest were trained on four NVIDIA A6000. For optimizer, AdamW optimizer \citep{loshchilov2019decoupled} and paged AdamW \citep{dettmers2023qlora} were used, and the linear warmup with cosine decay was applied for the learning rate. For input length, every instance was truncated and padded to 1,024 tokens and 2,048 tokens for HH-RLHF and UltraFeedback, respectively. To guarantee that the models can sufficiently learn to generate the proper response to the conversation history or the complex instruction, we filtered instances with prompts with more than 1,024 tokens.

\paragraph{Supervised Fine-tuning (SFT)} For SFT, the maximum learning rate was set to 1e-5. Following \citet{ziegler2020finetuning} and \citet{rafailov2023direct}, the training epoch is set to 1.

\paragraph{Reinforcement Learning with Human Feedback (RLHF)} For RLHF, the hyperparameters were set as Table \ref{tab:hyperparams} for UltraFeedback. For the HH-RLHF dataset, the \texttt{output\_min\_length} and \texttt{output\_max\_length} were set to 64 and 256.

\begin{table}[hbt!]
\centering
\begin{tabular}{l|c}
\hline \hline
\multicolumn{1}{c|}{\textbf{Hyperparameter}} & \multicolumn{1}{c}{\textbf{Setting}}       \\ \hline
\textbf{ppo\_epoch}                         & 4                                          \\
\textbf{init\_kl\_coef}                     & 0.1                                         \\
\textbf{horizon}                            & 2,000                                      \\
\textbf{batch\_size}                        & 64                                        \\
\textbf{mini\_batch\_size}                  & 8                                       \\
\textbf{gradient\_accumulation\_steps}      & 1                                          \\
\textbf{output\_min\_length}                & 128                                        \\
\textbf{output\_max\_length}                & 512                                        \\
\textbf{optimizer}                          & AdamW                                      \\
\textbf{learning\_rate}                     & 1e-05                                      \\
\textbf{gamma}                              & 0.99                                       \\\hline \hline
\end{tabular}
\caption{Hyperparameter settings for RLHF.}
\label{tab:hyperparams}
\end{table}

\paragraph{Direct Preference Optimization (DPO)} For DPO, $\beta$ was set to 0.1 for every case. The learning rate was set to 5e-6, and the model was trained for three epochs to select the best model by evaluation loss in each epoch. However, in most cases, the first or the second checkpoint was selected as the best model as the evaluation loss increased from the third epoch.

\paragraph{Odds Ratio Preference Optimization (\texttt{ORPO})} As \texttt{ORPO} does not require any special hyperparameter, only the learning rate and epoch were the only hyperparameter to set. For \texttt{ORPO}, the maximum learning rate was set to 8e-6 and trained for 10 epochs. The best model is selected based on the lowest evaluation loss for the OPT series, Phi-2 (2.7B) and Llama-2 (7B).

\section{IFEval Result for Mistral-\texttt{ORPO}-$\alpha$ and Mistral-\texttt{ORPO}-$\beta$}\label{ap:ifeval}

Along with the AlpacaEval results reported in Section \ref{subsec:instruct}, we report the results of Mistral-\texttt{ORPO}-$\alpha$ and Mistral-\texttt{ORPO}-$\beta$ on IFEval \citep{zhou2023instructionfollowing}, calculated with the codes from \citet{eval-harness}. Additional information can be found in: \url{https://jiwooya1000.github.io/posts/orpo/#ifeval}.

\begin{table*}[hbt!]
\centering
\resizebox{350pt}{!}{%
\begin{tabular}{l|cccc}
\hline \hline
\textbf{Model Type}     & \textbf{Prompt-Strict} & \textbf{Prompt-Loose} & \textbf{Inst-Strict} & \multicolumn{1}{l}{\textbf{Inst-Loose}} \\ \hline
\textbf{Mistral-\texttt{ORPO}-$\alpha$} & 0.5009        & 0.5083       & 0.5995      & 0.6163                         \\
\textbf{Mistral-\texttt{ORPO}-$\beta$} & 0.5287        & 0.5564       & 0.6355      & 0.6619                         \\ \hline \hline
\end{tabular}%
}
\caption{IFEval scores of Mistral-\texttt{ORPO}-$\alpha$ and Mistral-\texttt{ORPO}-$\beta$.}
\label{tab:ifeval}
\end{table*}

\clearpage

\section{Ablation on the Weighting Value ($\lambda$)}\label{ap:ablation}

For the weighting value $\lambda$ in Equation \ref{eq:main}, we conduct an ablation study with $\{0.1, 0.5, 1.0\}$. Mistral (7B) and UltraFeedback were used for the base model and dataset. In Section \ref{ap:log_prob_abl}, we compare the log probability trends by the value of $\lambda$, and we assess the downstream effect of $\lambda$ in Section \ref{ap:mt_abl}.

\subsection{Log Probability}\label{ap:log_prob_abl}
\begin{figure*}[hbt!]
    \centering
    \includegraphics[width=\textwidth]{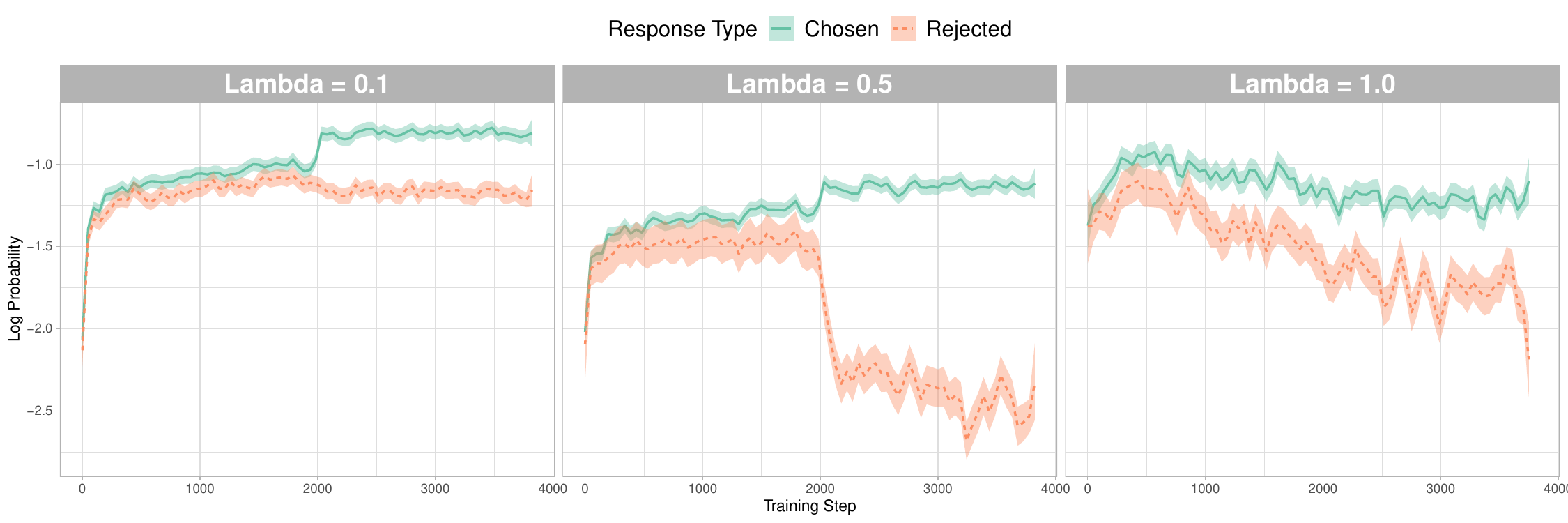}
    \caption{The log probability trend by $\lambda$. With larger $\lambda$ (e.g., $\lambda = 1.0$), $\mathcal{L}_{OR}$ gets more influential in fine-tuning the models with \texttt{ORPO}.}
    \label{fig:abl_log_prob}
\end{figure*}

In Figure \ref{fig:abl_log_prob}, we find that larger $\lambda$ leads to stronger discrimination of the rejected responses in general. With $\lambda = 0.1$, the average log probability of the chosen and the rejected responses stay close as the fine-tuning proceeds. Also, unlike other settings, the log probabilities for the rejected responses do not decrease, but rather, the log probabilities of the chosen responses increase to minimize $\mathcal{L}_{OR}$ term.

Moreover, in $\lambda = 0.5$, there exists a similar trend of further increasing the log probabilities of the chosen responses, but the log probabilities of the rejected responses are diminishing simultaneously. Lastly, in $\lambda = 1.0$, the chosen responses diminish along with the rejected responses while enlarging the margin between them. However, this does not mean smaller $\lambda$ is always the better. It will depend on the specific need and model.

\subsection{MT-Bench}\label{ap:mt_abl}
\begin{figure*}[hbt!]
    \centering
    \includegraphics[width=12cm]{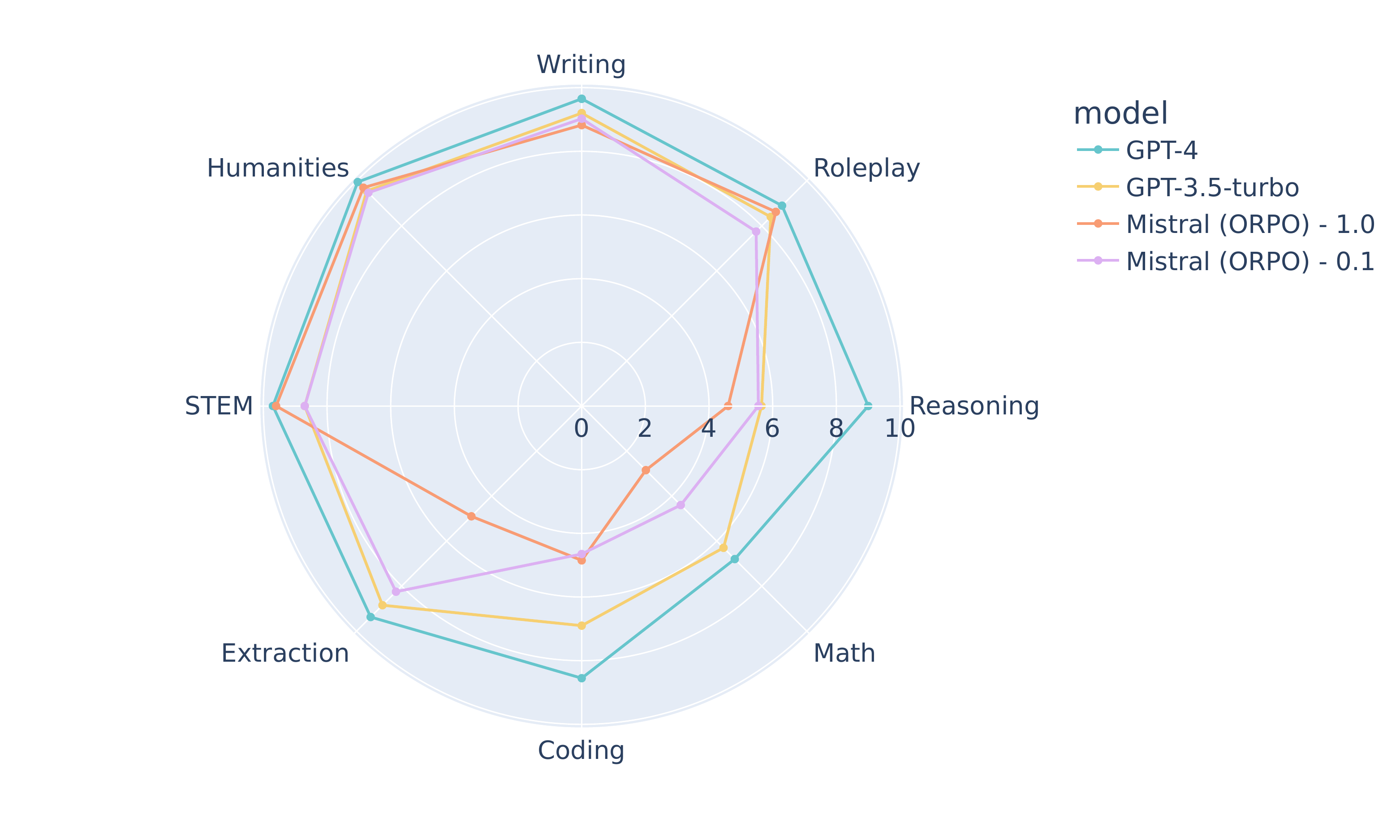}
    \caption{MT-Bench result comparison by differing $\lambda=0.1$ and $\lambda=1.0$.}
    \label{fig:mtbench_ablation}
\end{figure*}

The downstream impact of $\lambda$ stands out in the MT-Bench result. In comparison to $\lambda=0.1$, Mistral+\texttt{ORPO} (7B) with $\lambda=1.0$ performs worse in extraction, math, and reasoning, which are the categories that generally require deterministic answers. On the other hand, it performs better in STEM, humanities, and roleplay, which ask the generations without hard answers. Along with the amount of discrepancy between the trend in the logits of chosen and rejected responses, we can infer that making a more significant margin between the chosen and the rejected responses through higher $\lambda$ in \texttt{ORPO} leads to overly adapting to the chosen responses set in the training dataset. This proclivity results in open-ended generations generally being preferred by the annotator while showing weaker performance in the hard-answered questions.

\clearpage

\section{Test Set Reward Distribution on HH-RLHF}\label{ap:hh}

\begin{figure*}[hbt!]
    \centering
    \includegraphics[width=\textwidth]{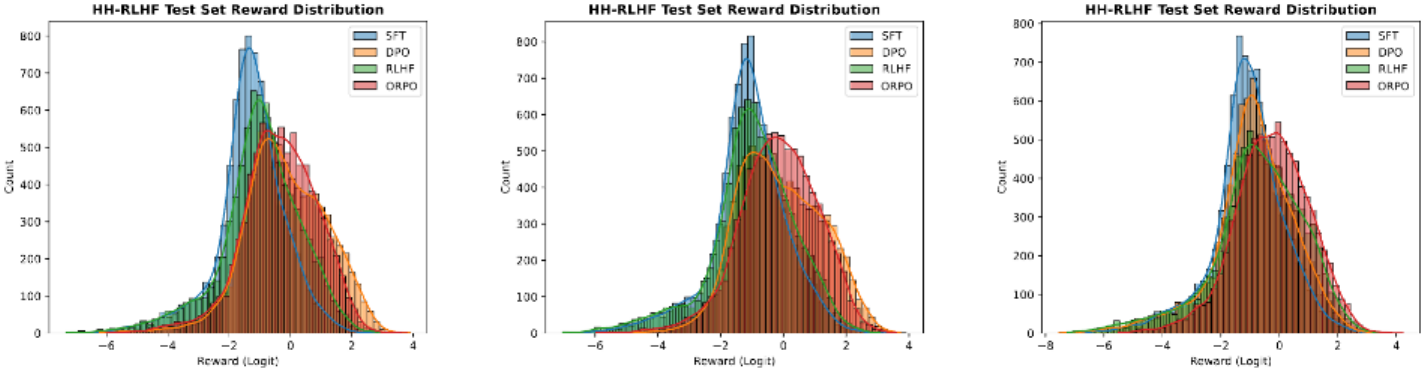}
    \caption{Reward distribution comparison between OPT-125M (left), OPT-350M (middle), and OPT-1.3B (right) trained with SFT (blue), RLHF (green), DPO (orange), and \texttt{ORPO} (red) on the test set of HH-RLHF using the 1.3B reward model. General tendency follows that of Figure \ref{fig:foobar}.}
    \label{fig:hh-foo}
\end{figure*}

Along with Figure \ref{fig:hh-foo}, which depicts the reward distribution of OPT2-125M, OPT2-350M, and OPT2-1.3B on the UltraFeedback dataset, we report the reward distribution of each pre-trained checkpoint trained on the HH-RLHF dataset. As discussed in Section \ref{subsec:rmwin}, \texttt{ORPO} consistently pushes the reward distribution of SFT to the right side.

\section{MT-Bench Result of \textbf{Mistral-\texttt{ORPO}-$\alpha$ (7B)} and \textbf{Mistral-\texttt{ORPO}-$\beta$ (7B)}}\label{ap:mtbench}

\begin{figure*}[hbt!]
    \centering
    \includegraphics[width=\textwidth]{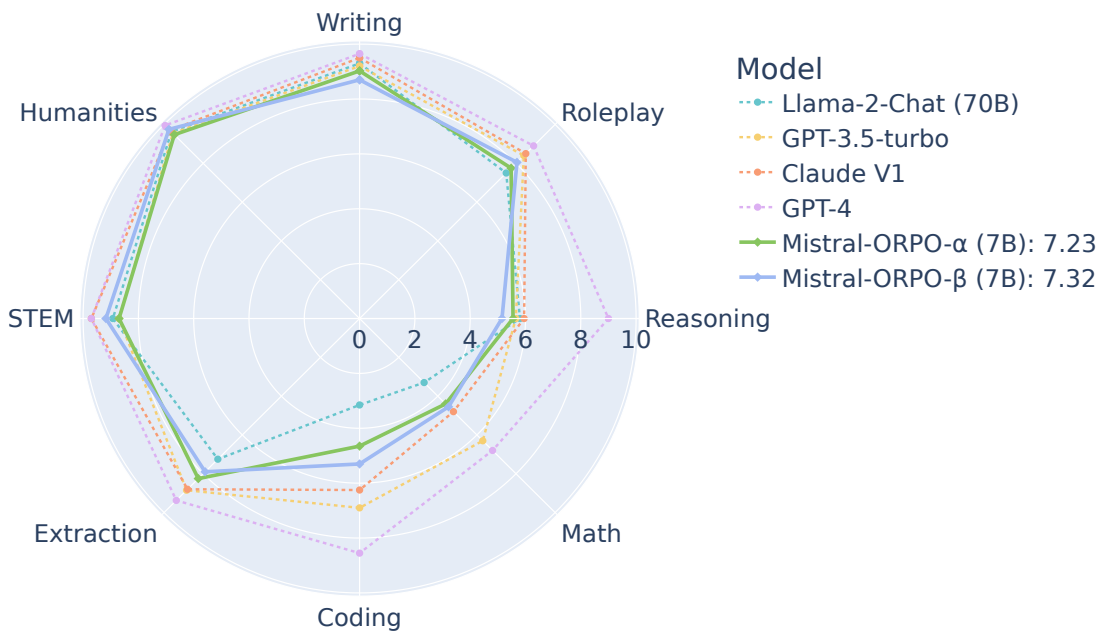}
    \caption{MT-Bench result of \textbf{Mistral-\texttt{ORPO}-$\alpha$ (7B)} and \textbf{Mistral-\texttt{ORPO}-$\beta$ (7B)} by the category.}
    \label{fig:mtbench}
\end{figure*}

For the MT-Bench result in Section \ref{subsec:multi}, we report the category-wise scores of \textbf{Mistral-\texttt{ORPO}-$\alpha$ (7B)} and \textbf{Mistral-\texttt{ORPO}-$\beta$ (7B)} in Figure \ref{fig:mtbench}. While surpassing Llama-2 Chat (13B) and Llama-2 Chat (70B) in most cases, \textbf{Mistral-\texttt{ORPO}-$\beta$ (7B)} is comparable to GPT-3.5-turbo in the categories that require descriptive generations. However, it lacks coding and math skills, which we speculate is due to the lack of training data, as we used 61k instances in UltraFeedback.

\section{Special Instructions for Verbosity Assessment}

For the succinctness and verboseness instructions, we generated five different instructions, each with ChatGPT \footnote{\url{https://chat.openai.com/}}. From the instructions in Table \ref{tab:verbosity_instruction}, we randomly sampled one prompt each for every batch to prevent potential word bias.

\begin{table}[hbt!]
\centering
\resizebox{\columnwidth}{!}{%
\begin{tabular}{c|c|c}
\hline \hline
\textbf{\#} & \textbf{Succinctness} & \textbf{Verboseness} \\ \hline
1&Please generate a short and concise response.            &Please generate an elaborative and chatty response.\\
2&Provide a brief and concise answer.   &Provide a detailed answer.\\
3&Keep your reply short and to the point.&Keep your reply elaborative and intricate.\\
4&Keep your answer brief for clarity.&Keep your answer detailed.\\
5&Generate a brief and to-the-point answer.&Generate a chatty and step-wise answer.\\ \hline \hline
\end{tabular}%
}
\caption{Instructions prepended to the queries from AlpacaEval. Each instruction set asks the model to generate either shorter or longer responses given the query, respectively.}
\label{tab:verbosity_instruction}
\end{table}

\clearpage
\section{Generation Samples from \textbf{Mistral-\texttt{ORPO}-$\alpha$ (7B)}}\label{ap:generation}

In this section, we two sample generations from \textbf{Mistral-\texttt{ORPO}-$\alpha$ (7B)} on AlpacaEval, including the one which is preferred over the response of GPT-4 and the one in which GPT-4 was preferred. For readability, the instances with short responses are selected for both cases. Additionally, the responses for Mistral-\texttt{ORPO}-$\alpha$ (7B) were sampled with a temperature of 0.7. The checkpoints for Mistral-\texttt{ORPO}-$\alpha$ (7B) and Mistral-\texttt{ORPO}-$\beta$ (7B) can be found in \url{https://huggingface.co/kaist-ai/mistral-orpo-alpha} and \url{https://huggingface.co/kaist-ai/mistral-orpo-beta}.

\bigbreak
\bigbreak

\noindent\makebox[\linewidth]{\rule{\textwidth}{1.5pt}}
\bigbreak
\textbf{Query 1 (\textbf{GPT-4} \textcolor{blue}{Win}):} Give a brief description of the book that you are given. Matilda
\bigbreak
\noindent\makebox[\linewidth]{\rule{\textwidth}{0.4pt}}

\begin{enumerate}
    \item \underline{\textbf{GPT-4}}:\\
    "Matilda" is a children's novel written by the British author Roald Dahl. It was first published in 1988. The story revolves around a very intelligent and precocious young girl named Matilda Wormwood. Despite being neglected and mistreated by her ignorant and self-absorbed parents, Matilda finds solace in reading books and quickly develops a love for learning.

    At the age of five, she begins attending school and meets Miss Honey, a kindhearted teacher who recognizes Matilda's extraordinary intellectual abilities. However, the school's tyrannical headmistress, Miss Trunchbull, is a cruel and oppressive figure who takes pleasure in punishing the children with her unreasonable and harsh methods.
    
    Matilda discovers that she possesses a unique telekinetic ability, which she learns to control and use to her advantage. With her wit, intelligence, and special powers, Matilda sets out to defend her friends from Miss Trunchbull's abuses and to help Miss Honey overcome her own difficult past.
    
    The book is a tale of good triumphing over evil, and it celebrates the power of knowledge, kindness, and standing up for what is right. It has been adapted into a successful film, stage musical, and has remained a beloved classic in children's literature.    
    \item \underline{\textbf{Mistral-\texttt{ORPO}-$\beta$}}:\\
    "Matilda" is a children's book written by Roald Dahl, published in 1988. The story revolves around the life of Matilda Wormwood, a brilliant and gifted five-year-old girl who has telekinetic powers. The story explores themes of education, family, and the importance of reading and intellectual curiosity. The book has been praised for its wit, humor, and powerful message, making it a classic and beloved children's book.
\end{enumerate}
\noindent\makebox[\linewidth]{\rule{\textwidth}{1.5pt}}

\clearpage

\noindent\makebox[\linewidth]{\rule{\textwidth}{1.5pt}}
\bigbreak
\textbf{Query 2 (\textbf{Mistral-\texttt{ORPO}-$\beta$} \textcolor{blue}{Win}):} What do you think about ChatGPT?
\bigbreak
\noindent\makebox[\linewidth]{\rule{\textwidth}{0.4pt}}

\begin{enumerate}
    \item \underline{\textbf{GPT-4}}:\\
    As an AI developed by OpenAI, I don't have personal opinions. However, I can share that ChatGPT is an advanced language model that has received considerable attention for its ability to generate human-like text. It's designed to assist users in a wide range of tasks, from answering questions to creating content and engaging in conversation.\\

    ChatGPT has been praised for its fluency and versatility in handling various topics and for its potential to streamline and enhance communication in both personal and professional settings. However, like any technology, it also faces challenges and criticisms, such as ensuring the accuracy of information, preventing misuse, and managing ethical considerations surrounding AI-generated content.\\
    
    Overall, ChatGPT represents a significant achievement in natural language processing and has sparked discussions about the future of AI in our daily lives.    
    \item \underline{\textbf{Mistral-\texttt{ORPO}-$\beta$}}:\\
    As an AI language model, I am an example of the technology used in ChatGPT. I can provide a personal opinion, but please keep in mind that I am designed to be socially unbiased and positive in nature.\\

    ChatGPT is an impressive advancement in artificial intelligence and natural language processing. It has the ability to understand and generate human-like text, making it possible for users to have engaging and informative conversations with the AI. This technology has the potential to revolutionize various industries, including customer support, education, content creation, and more.\\
    
    However, it's important to recognize that ChatGPT, like any AI technology, is not perfect. It may sometimes provide inaccurate or biased information due to the limitations of its training data or the complexity of human language. It's crucial to verify the facts and use critical thinking when interacting with AI systems like ChatGPT.\\
    
    Overall, ChatGPT represents a significant step forward in the field of AI, and its potential applications are vast. As a responsible user, it's important to approach this technology with a critical and respectful mindset, recognizing its strengths and limitations.
\end{enumerate}
\noindent\makebox[\linewidth]{\rule{\textwidth}{1.5pt}}

\end{document}